\documentclass[hidelinks]{article}

\usepackage[final, nonatbib]{neurips_2020}

\usepackage[utf8]{inputenc} 
\usepackage[T1]{fontenc}    
\usepackage{hyperref}       
\usepackage{url}            
\usepackage{booktabs}       
\usepackage{amsfonts}       
\usepackage{nicefrac}       
\usepackage{microtype}      
\usepackage{enumitem}       
\usepackage{subcaption}
\usepackage{graphicx}
\graphicspath{ {./images/} }
\usepackage{wrapfig}
\usepackage{xcolor}
\usepackage{multirow}
\usepackage{amsmath}
\usepackage{float}
\numberwithin{equation}{section}
\definecolor{bleudefrance}{rgb}{0.19, 0.55, 0.91}

\title{Forecasting Black Sigatoka Infection Risks with Latent Neural ODEs}

\author{%
  Yuchen Wang\quad \quad Matthieu Chan Chee \quad \quad Ziyad Edher  \\
  \textbf{Minh Duc Hoang \quad\quad Shion Fujimori}\\
  \textbf{Sornnujah Kathirgamanathan \quad\quad Jesse Bettencourt} \\
  University of Toronto \\
  \{raina.wang, matthieu.chanchee, ziyad.edher, minhduc.hoang, \\
  shion.fujimori, sornnujah.kathirgamanathan\}@mail.utoronto.ca, \\
  jessebett@cs.toronto.edu
}

\begin{document}

\maketitle
\begin{abstract}
Black Sigatoka disease severely decreases global banana production, and climate change aggravates the problem by altering fungal species distributions. Due to the heavy financial burden of managing this infectious disease, farmers in developing countries face significant banana crop losses. Though scientists have produced mathematical models of infectious diseases, adapting these models to incorporate climate effects is difficult. We present MR. NODE (Multiple predictoR Neural ODE), a neural network that models the dynamics of black Sigatoka infection learnt directly from data via Neural Ordinary Differential Equations. Our method encodes external predictor factors into the latent space in addition to the variable that we infer, and it can also predict the infection risk at an arbitrary point in time. Empirically, we demonstrate on historical climate data that our method has superior generalization performance on time points up to one month in the future and unseen irregularities. We believe that our method can be a useful tool to control the spread of black Sigatoka. \footnote{Our code and datasets used are available at \url{https://github.com/UofTrees/ProjectX2020} and \url {https://drive.google.com/drive/folders/1VhUxAUcGQwrA98b-BtE18IBWkm5n4bWP?usp=sharing} respectively.}
\end{abstract}

\section{Introduction}

As human activities continue to influence the climate, the range of optimal weather conditions for many plant pathogens grows. Caused by the fungal plant pathogen \textit{Mycosphaerella fijiensis}, the black Sigatoka disease is an economically significant example of this phenomenon. Research has proven this leaf-spot infectious disease’s detriment to banana and plantain crops globally. The Food and Agriculture Organization of the United Nations reported that between 2007 and 2009, St. Vincent and the Grenadines faced a 90\% decline in banana crop production due to this disease \cite{fao_2013}. Managing the impact of \textit{M. fijiensis} infection can decrease the cost of crop production by up to approximately 25\%. However, many local farmers in developing countries have no access to disease management facilities due to financial boundaries \cite{fao_2013}. Thus, predicting upcoming \textit{M. fijiensis} infections on banana plants would allow farmers to take appropriate preventative measures and mitigate disease management costs. Though attempted by \cite{ochoa_2016, calvo-valverde_2017}, powerful models for forecasting infections of black Sigatoka have not yet been developed.

Advances in machine learning algorithms such as recurrent neural networks (RNN) and long short-term memories (LSTMs) \cite{hochreiter_1997} tackled time series modelling. However, these models have difficulty modelling irregularity in time. Recently, \cite{chen_2018} addressed this limitation by learning a differential equation parameterized as a neural network directly from data. Using this neural ordinary equation (Neural ODE) as a key component, \cite{chen_2018} developed new models for time-series forecasting tasks that can incorporate data arriving at arbitrary times. In this study, we predict the spread of black Sigatoka based on varying microclimatic conditions by adopting the latent neural ODE approach.

Importantly, our model maintains high predictive capabilities even when provided with irregularly-sampled data. This is particularly pertinent as missing values frequently appear in agriculture datasets recorded in developing countries.
\paragraph{Summary of Contributions}
\begin{itemize}
    \item We propose Multiple predictoR Neural ODE, a type of ODE-Net that defines a latent generative function. Our method incorporates a look-up function in the ODE dynamics, which concatenates the to-be-inferred variable with its multiple predictors. By also outputting only the variable of interest, we extended the architecture of latent Neural ODEs \cite{chen_2018} to model and infer multivariate time series with external factors. See section (\ref{Multiple predictoR Neural ODE}).
    \item We demonstrate our method's effectiveness in learning the dynamics of the generated black Sigatoka disease data based on the historical climate database from the Japanese Meteorological Agency 55-Year reanalysis (JRA-55) \cite{kobayashi_2015}.
    \item  In addition, we trained and evaluated RNNs and LSTMs on our dataset as baseline models and demonstrate that our method outperforms them. See section (\ref{experiments}).
\end{itemize}

\section{Background}
\subsection{Epidemic modelling}
Infectious diseases have been studied by many mathematical models. In 1925, \cite{mkendrick_1925} introduced a differential equation model for epidemics that considers a fixed population with susceptible, infected and recovered individuals (SIR). In recent years, \cite{ochoa_2016} suggested modeling black Sigatoka using the Auto Regressive Integrated Moving Average (ARIMA) model, and \cite{calvo-valverde_2017} applied several machine learning techniques including recurrent neural networks in forecasting the development rate of the black Sigatoka disease. However, no work predicts disease risks with state-of-the-art sequential methods.

\subsubsection{A Statistical Model} \label{A statistical model}
For a number of hypothetical cohorts of \textit{M. fijiensis} spores, \cite{bebber_2019} modelled the infection of black Sigatoka as a probabilistic survival process depending on three microclimatic condition variables:  relative humidity (RH), canopy temperature (T), and moisture storage on canopy (CM). A cohort of spores germinates and infects its host during wet periods and ceases the process during dry ones. A wet period is a succession of at least three contiguous time points whereby CM $> 0$ meters or RH > $98\%$.

\begin{equation}\label{eq:1}
r(T) = \left(\frac{T_{max}-T}{T_{max}-T_{opt}}\right)\left(\frac{T-T_{min}}{T_{opt}-T_{min}}\right)^{\frac{T_{opt}-T_{min}}{T_{max}-T_{opt}}}
\end{equation}
\begin{equation}\label{eq:2}
H(t, T) = r(T)\left(\frac{t}{\alpha}\right)^\gamma 
\end{equation}
\begin{equation}\label{eq:3}
    F(t, T) = 1 - e^{-H(t, T)} 
\end{equation}
\begin{equation}\label{eq:4}
    Y(t, T) = \beta F(t, T)
\end{equation}

Given estimated cardinal temperatures (the minimum $T_{min}$, the optimum $T_{opt}$ and the maximum $T_{max}$, in degree Celsius), (Equation \ref{eq:1}) can be used to determine a relative rate $r$ for spore growth. With the scale factor $\alpha$ and the shape parameter $\gamma$ we then calculate a cumulative Weibull hazard function $H$ at each time point $t$ in a wet period (Equation \ref{eq:2}). Via (Equation \ref{eq:3}), $H$ further determines $F$, the fraction of a cohort of spores that has infected a leaf. Finally, $Y$, the number of cohorts of \textit{M. fijiensis} spores that caused infection, is computed as the product of $F$ with the number of cohorts $\beta$ (Equation \ref{eq:4}). Thus, the infection risk is defined as the sum of hourly spore cohorts that infect a leaf over a time interval.

\subsection{Machine Learning Methods for Time Series}
A recurrent neural network (RNN) is a class of artificial neural networks for sequential modelling. Internal states in an RNN connect to each other in a temporal order, enabling the network to process inputs of variable lengths. \cite{hochreiter_1997} invented long short-term memory (LSTM), which reinforced the RNN architecture by solving its "gradient vanishing problem". RNNs and LSTMs are known to perform well on tasks such as language modelling, whereby data is sampled at regular intervals \cite{lamb_2016}. However, as suggested by \cite{chen_2018}, applying RNNs to irregularly-sampled data can be challenging. Such data is typically discretized into bins of fixed duration, thus leading to complications if missing data exists.

\subsection{Latent Neural ODEs}
\begin{figure}[htp]
    \centering
    \includegraphics[width=1\textwidth]{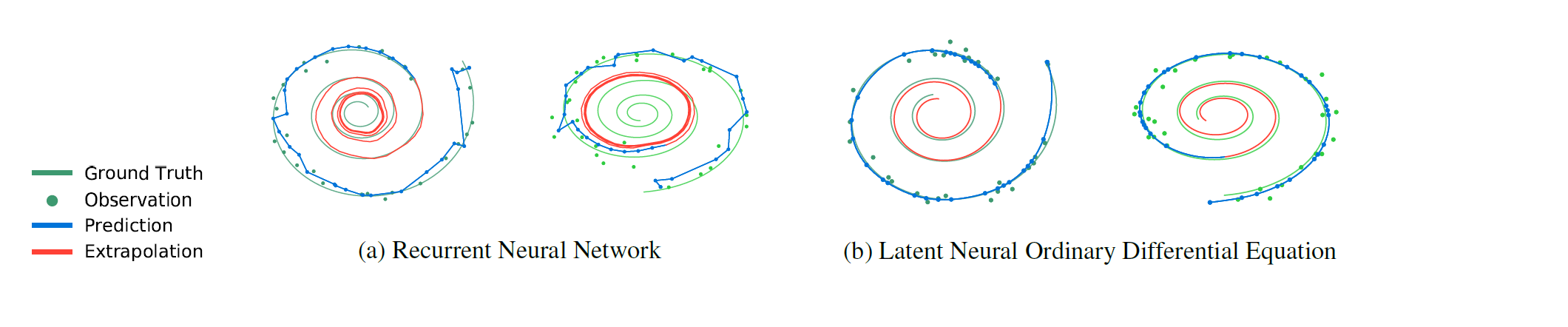}
    \caption{Comparison between RNNs and latent Neural ODE in \cite{chen_2018}. The latent Neural ODE outperforms the RNN when given irregularly-sampled data.}
    \label{fig:dataset}
\end{figure}

\cite{chen_2018} introduced Neural ODEs, a family of deep neural networks that parameterize the derivative of the hidden state using a neural network, which is fed into a black-box ODE solver. Latent Neural ODEs adopt Neural ODE as a critical part of the model to generatively model continuous time series. \cite{chen_2018} demonstrated that latent Neural ODEs could outperform RNNs in terms of extrapolation as well as modelling irregularities. Their approach is as follows:
\begin{enumerate}[label=(\roman*)]
\item Assume that the given time series can be represented by a latent trajectory uniquely defined by an initial hidden state $z_{t_0}$ and a time-invariant dynamics function $f = \frac{dz}{dt}$. $f$ is parameterized by a feed-forward neural network.
\item An encoder RNN takes in data $x_{t_0}, ..., x_{t_N}$ for observed time steps $t_0, ..., t_N$ and produces the parameters $\mu$ and $\sigma$ for a Gaussian posterior over the initial state $z_{t_0}$ in latent space: 
$$q(z_{t_0}|\{x_{t_i}, t_i\}_i, \phi) = \mathcal{N}(z_{t_0}|\mu, \sigma)$$
\item Sample $z_{t_0} \sim q(z_{t_0}|\{x_{t_i}, t_i\}_i, \phi)$
\item The initial state $z_{t_0}$, dynamics function $f$, and the time steps for prediction and extrapolation $t_0, ..., t_N$, $t_{N+1}, ..., t_M$ are fed into a black-box ODE solver. This ODE solver applies techniques such as the Euler method or the Dormand-Prince method \cite{dormand_1980} to generate values $z_{t_0}, ..., z_{t_N}, z_{t_{N+1}}, ..., z_{t_M}$.
\item A decoding neural-net maps the latent space values $z_{t_0}, ..., z_{t_N}, z_{t_{N+1}}, ..., z_{t_M}$ back to data space, thus giving $\hat{x}_{t_0}, ..., \hat{x}_{t_N}, \hat{x}_{t_{N+1}}, ..., \hat{x}_{t_M}$.
\end{enumerate}

\section{Datasets}
\begin{figure}[H]
\centering
\includegraphics[width=0.7\textwidth]{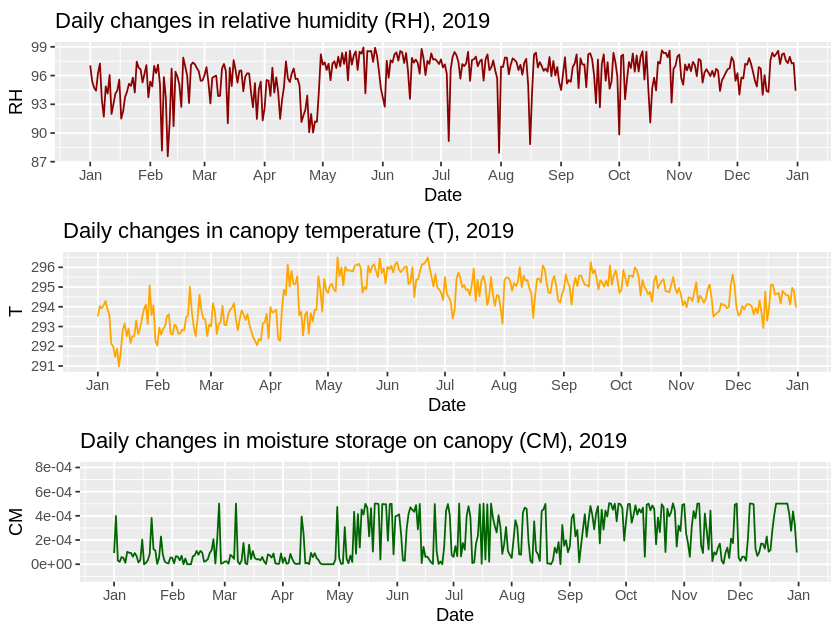}
\caption{The progression of RH, T, and CM throughout 2019 in Costa Rica (83.812 W, 10.39 N)}\label{wrap-fig:1}
\end{figure} 
Since binding agreements between farmers and companies highly privatize crop disease datasets, we generated our own dataset using the disease risk model presented in \cite{bebber_2019} (see Section \ref{A statistical model}) from the Japanese Meteorological Agency 55-Year reanalysis (JRA-55) dataset \cite{kobayashi_2015}. JRA-55 comprises high spatio-temporal resolution climate data spanning different parts of the world, collected from 1958 to the present day. From the vast amount of longitude and latitude coordinates available in JRA-55, we chose to study Costa Rica and India since both are countries known for banana plantations. From the Spatial Production Allocation Model (SPAM) dataset \cite{spam_2010} of global production, we selected the longitude-latitude coordinates that have plentiful banana productions in 2010. For each of these coordinates, we obtained a 6-hourly multivariate time series for years 1958 - 2020 inclusive, which contains 91,556 time points and three microclimatic condition variables:  relative humidity (RH), canopy temperature (T), and moisture storage on canopy (CM). We plotted the three variables for year 2019 in Costa Rica (83.812 W, 10.39 N) in (Figure \ref{wrap-fig:1}).

\begin{figure}[htp]
    \begin{subfigure}[t]{0.6\textwidth}
    \includegraphics[width=\linewidth, height=4cm]{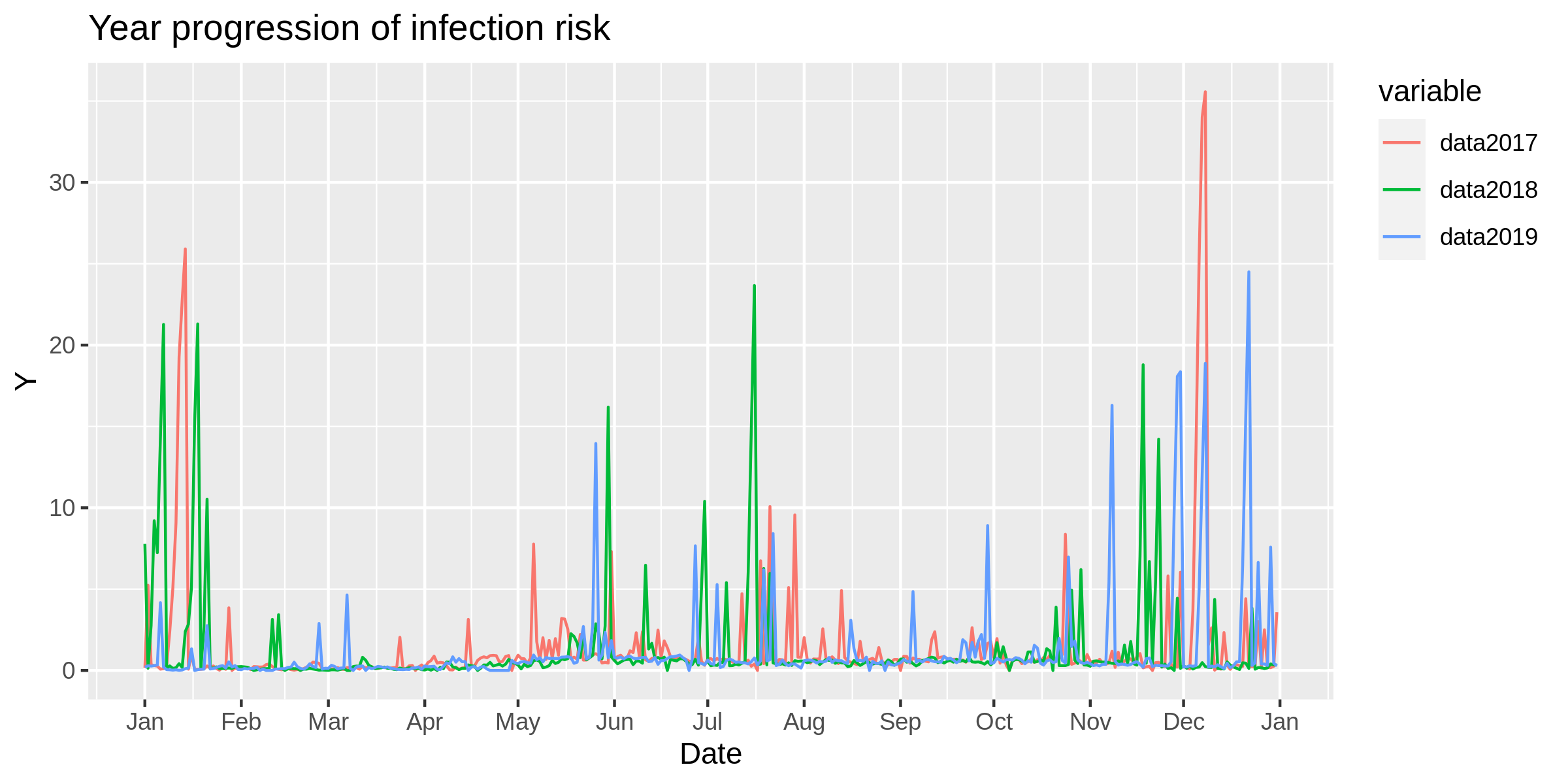}
    \caption{}
    \end{subfigure}
    \begin{subfigure}[t]{0.35\textwidth}
    \includegraphics[width=\linewidth]{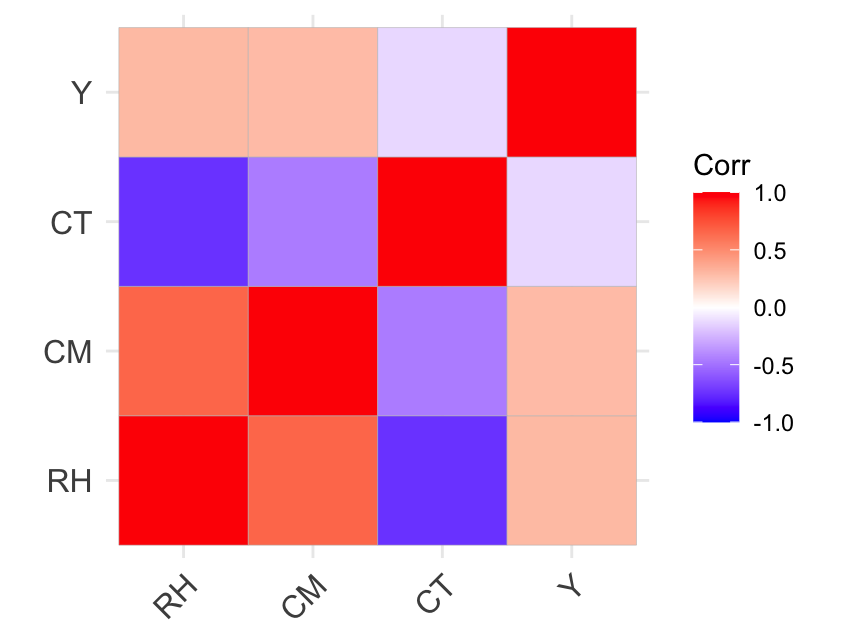}
    \caption{}
    \end{subfigure}
    \caption{Visualizations of our dataset. (a) The progression of the generated infection variable throughout 2017, 2018 and 2019 in Costa Rica (83.812 W, 10.39 N). (b) Correlation matrix of data variables in Costa Rica. RH has strong positive correlation with CM and strong negative correlation with CT. However, no signs show significant correlation between Y and any other variables. }
    \label{fig:data}
\end{figure}

To generate the infection variable $Y$ as laid out in Section \ref{A statistical model}, we utilized the best-fitting model parameters from the simulation experiments in \cite{bebber_2019}, where $T_{min} = 16.6$, $T_{opt} = 27.2$, $T_{max} = 30.3$, $\alpha = 32.6$, $\gamma = 1.76$ and $\beta = 37.6$. Hence we obtained a 6-hourly four-dimensional time series dataset (three microclimatic conditions and the infection variable). We considered our generated infection variable as the ``ground truth” in later experiments. (Figure \ref{fig:data}) shows the yearly comparison of the infection risk variable and correlation matrix among all variables.

\section{Methodology: Multiple predictoR Neural ODE \label{Multiple predictoR Neural ODE}}
\begin{figure}[H]
    \centering
    \begin{subfigure}[t]{0.87\textwidth}
    \includegraphics[width=0.87\textwidth]{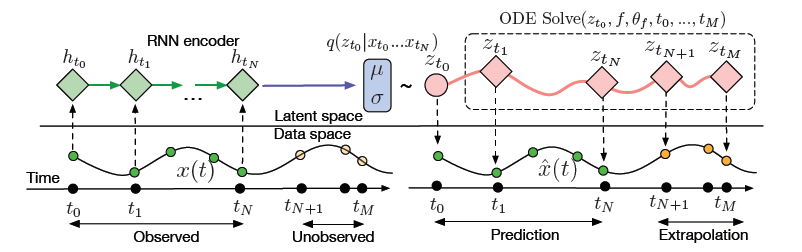}
    \caption{Computational graph of latent Neural ODE model presented by \cite{chen_2018}}
    \end{subfigure}
    \begin{subfigure}[t]{0.89\textwidth}
    \includegraphics[width=0.89\textwidth]{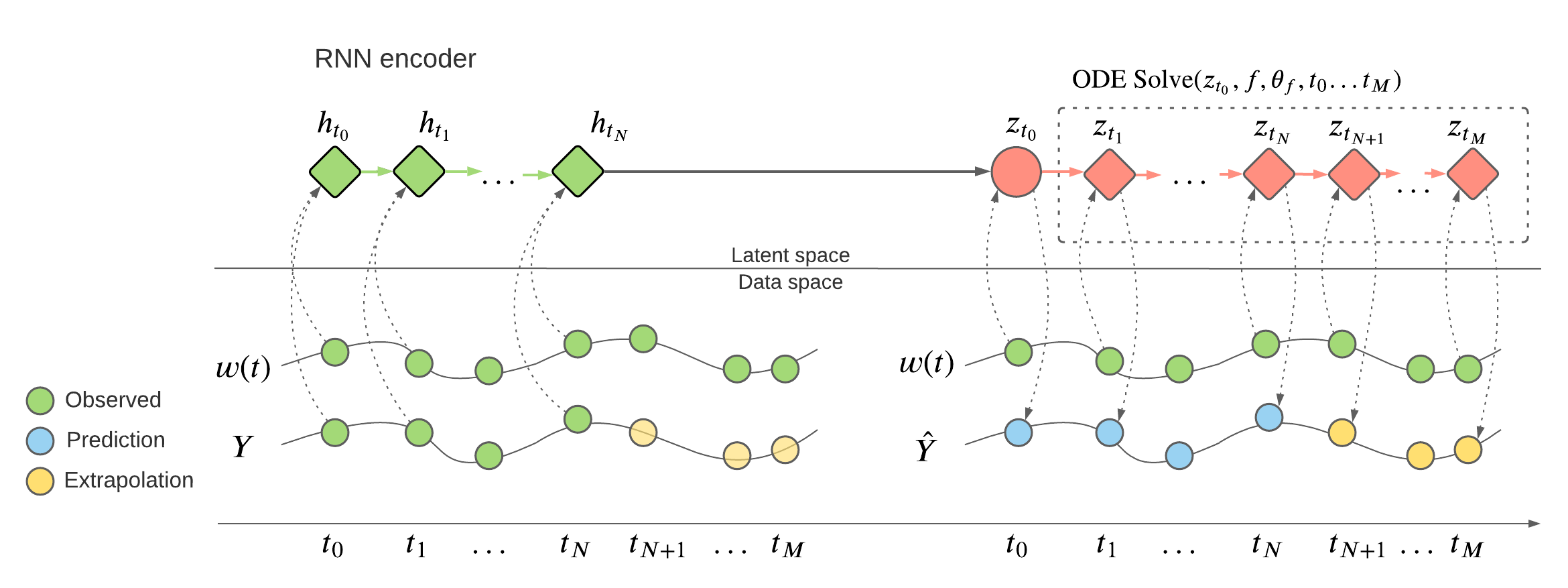}
    \caption{Computational graph of our model}
    \end{subfigure}
    \caption{Model architectures.}
    \label{fig:datavisual}
\end{figure}

We introduce the Multiple predictoR Neural ODE (MR. NODE), an architecture suitable to model time series data with external predictors. This is enabled by two key innovations. Firstly, we implemented a look-up function in the Neural ODE dynamics. A naive application of the latent Neural ODE system would learn latent dynamics of all the variables. However, this approach has high computation cost and departs from our goal to predict only the infection risks. Therefore, we discard the naive application and feed microclimate variables as external predictors directly into the latent space as given. In particular, we concatenate a continuous function $w(t)$ (the progression of external conditions through time $t$) to the encoded inputs, in the dynamics neural network $f$ before solving the ODEs. Secondly, instead of extrapolating the entire input space in the predictions, we train our latent Neural ODE system as a ``partial" autoencoder, which only outputs the disease variable $Y$.
\section{Experiments}\label{experiments}
In this section, we conducted a series of experiments to validate the following questions:
\begin{enumerate}
    \item Does our method extrapolate the number of cohorts that caused infection well into the future compared to baseline models? (long extrapolation)
    \item Does our method interpolate the number of infections well at unseen irregularities? (irregular interpolation)
\end{enumerate}
\subsection{Long Extrapolation}
For MR. NODE, we used an LSTM as the encoder and the Euler ODE solver in all experiments. In the training phase, MR. NODE encodes 128 6-hourly time points into the initial latent state and reconstructs the infection risk for those 128 time points. Loss is calculated using negative log likelihood of ground truths under Gaussian distributions with the predictions as means. In the validation phase, the model encodes $100$ time points. It then reconstructs the infection risk for those 100 time points and extrapolates for 150 further time points, which is equivalent to extrapolating 37.5 days into the future. We summarized the training, validation and test settings of MR. NODE in (Table \ref{table:setting}).

To simulate irregular time conditions, we randomly dropped a proportion ($p = \{0, 0.3, 0.5, 0.7\}$) of data points for each data window (Table \ref{table:setting}).  Notably, we performed data dropping for MR. NODE only during the testing phase because the model naturally learns a continuous dynamics throughout all the time points while training. On the other hand, we needed to retrain the baselines for every dropping rate and, as suggested in \cite{chen_2018}, each input is concatenated with the time difference since the previous time step.

\begin{figure}[H]
    \centering
    \begin{subfigure}{0.49\textwidth}
    \includegraphics[width=1\textwidth]{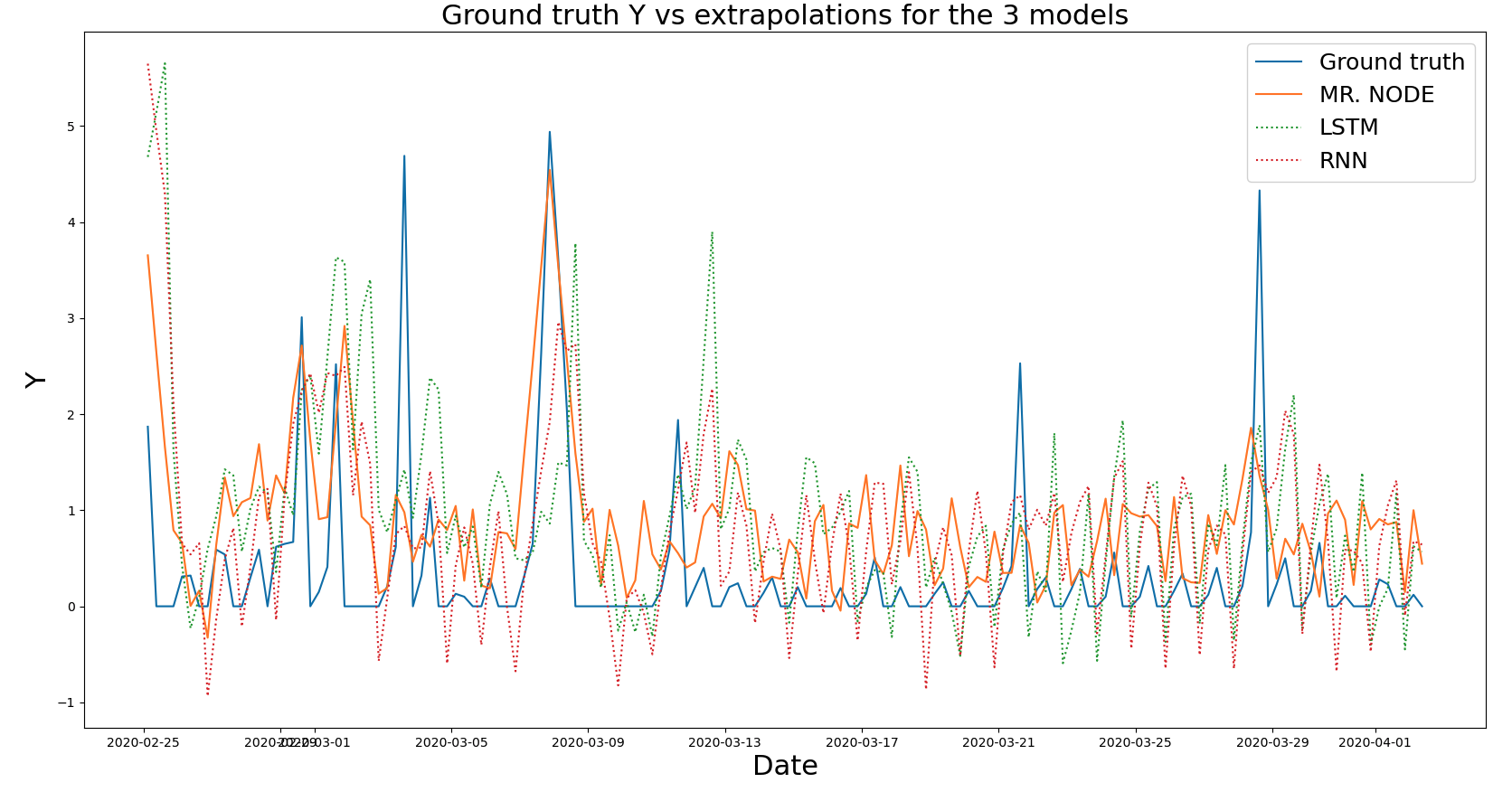}
    \caption{Feb. 25 2020 - Apr. 1 2020}
    \end{subfigure}
    \begin{subfigure}{0.49\textwidth}
    \includegraphics[width=1\textwidth]{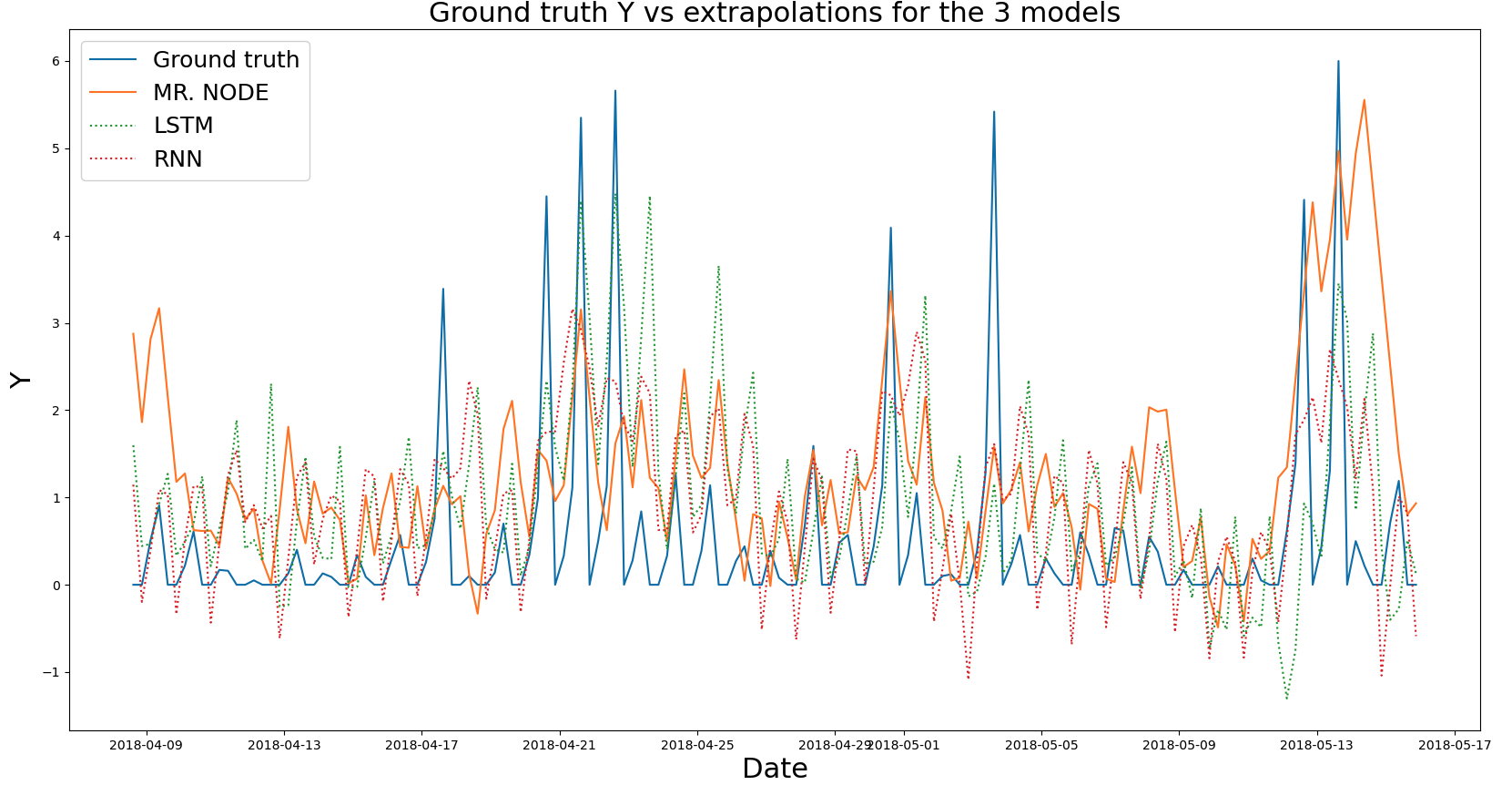}
    \caption{Apr. 9 2018 - May 17 2018}
    \end{subfigure}
    \caption{Extrapolated data windows for MR. NODE, RNN and LSTM trained and tested on Costa Rica data.}
    \label{fig:extra}
\end{figure}

\begin{table}
\centering
\begin{tabular}{|c|c|c|c|} 
\hline
& \# encoded  & \# reconstructed & \# extrapolated \\
\hline 
Training   & 128   & 128  & 0  \\
\hline 
Validation & 100 & 100  & 150  \\
\hline 
Extrapolation Test & 100/70/50/30 (drop rate 0/0.3/0.5/0.7) & 100  & 150  \\
\hline
Interpolation Test & 70/30/10 (drop rate 0.3/0.7/0.9) & 100 & 0  \\
\hline 
\end{tabular}
\vspace{1em}
\caption{Training, validation and test settings of MR. NODE}
\label{table:setting}
\end{table}
We compared our method with RNNs and LSTMs by calculating the mean square error (MSE) on extrapolated points in the testing phase. The results for our model trained on Costa Rica data are summarized in Table \ref{table:results1}. We also trained a model on combined data from both Costa Rica and India. Its individual performance on the test set from either Costa Rica or India is noted in Table \ref{table:results2}. The plots for extrapolated data windows in the testing phase against the ground truth can be found in (Figure \ref{fig:extra}). 

\begin{table}[H]
\centering
\begin{tabular}{|c|c|c|c}
\hline
Method & Drop rates & Test set - Costa Rica   \\
\hline 
\multirow{4}{4em}{RNN}   & 0  & 13.55  \\
& 0.3 & 13.51\\
& 0.5 & 12.62 \\
& 0.7 & 13.70 \\
\hline 
\multirow{4}{4em}{LSTM}   & 0  & 12.76  \\
& 0.3 & 12.71 \\
& 0.5 & 17.94 \\
& 0.7 & 14.12 \\
\hline 
\multirow{4}{4em}{MR. NODE}   & 0  & \textbf{12.16}  \\
& 0.3 & \textbf{12.29}\\
& 0.5 & \textbf{12.40} \\
& 0.7 & \textbf{12.68} \\
\hline
\end{tabular}
\vspace{1em}
\caption{Results (average MSE over the data windows) for models trained on Costa Rica data.}
\label{table:results1}
\end{table}

\begin{table}[H]
\centering
\begin{tabular}{|c|c|c|}
\hline
Method & Test set - Costa Rica & Test set - India  \\
\hline
RNN & 12.34 & 148.47 \\
\hline
LSTM & 12.43 & 133.89 \\
\hline
MR. NODE & \textbf{11.64} & \textbf{55.68} \\
\hline
\end{tabular}
\vspace{1em}
\caption{Results (average MSE over the data windows) for models trained on both Costa Rica and India data.}
\label{table:results2}
\end{table}

\subsection{Irregular Interpolation}
\begin{figure}[H]
    \centering
    \begin{subfigure}{0.49\textwidth}
    \includegraphics[width=1\textwidth]{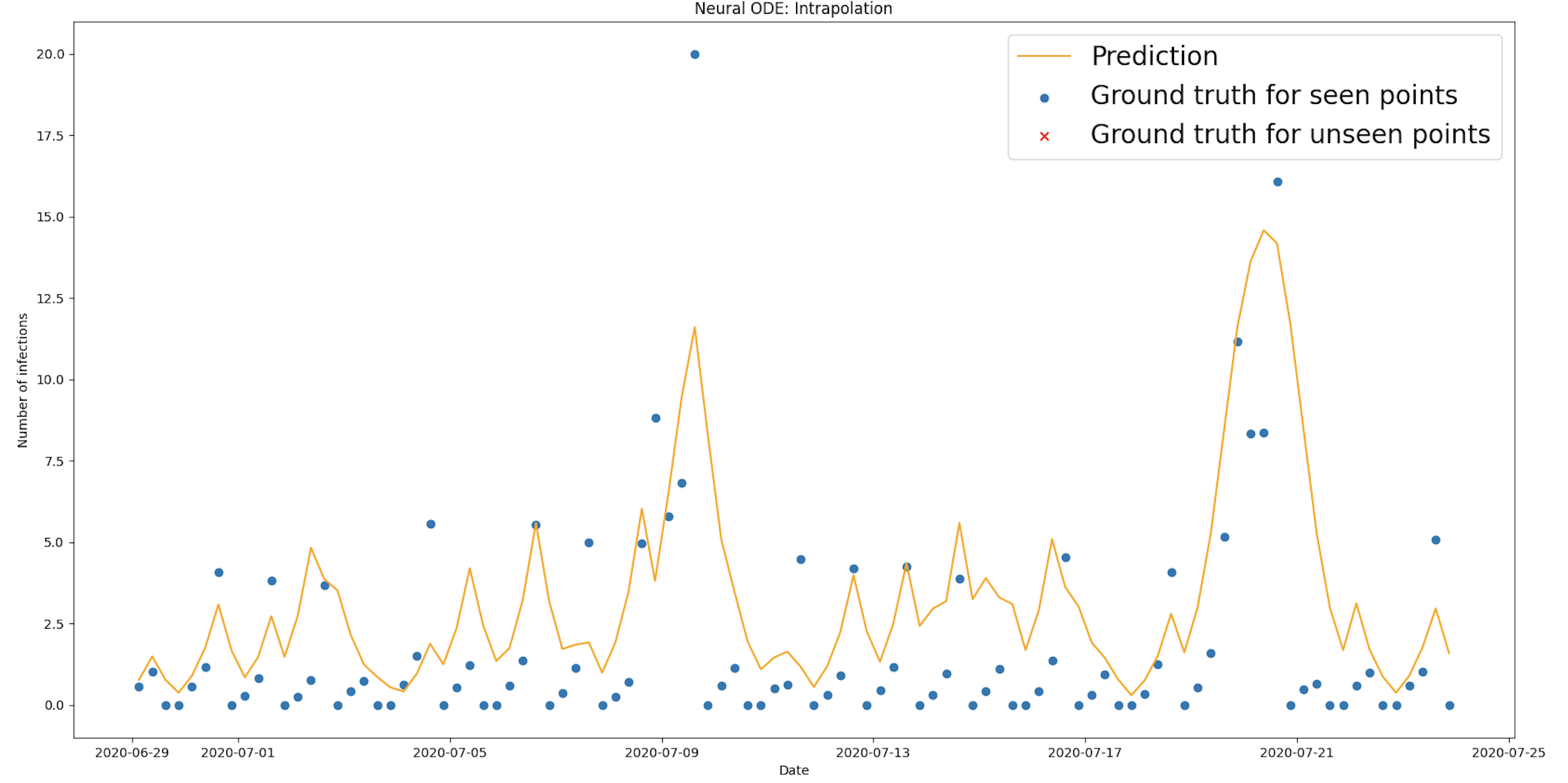}
    \caption{dropping rate $= 0$}
    \label{fig:0}
    \end{subfigure}
    \begin{subfigure}{0.49\textwidth}
    \includegraphics[width=1\textwidth]{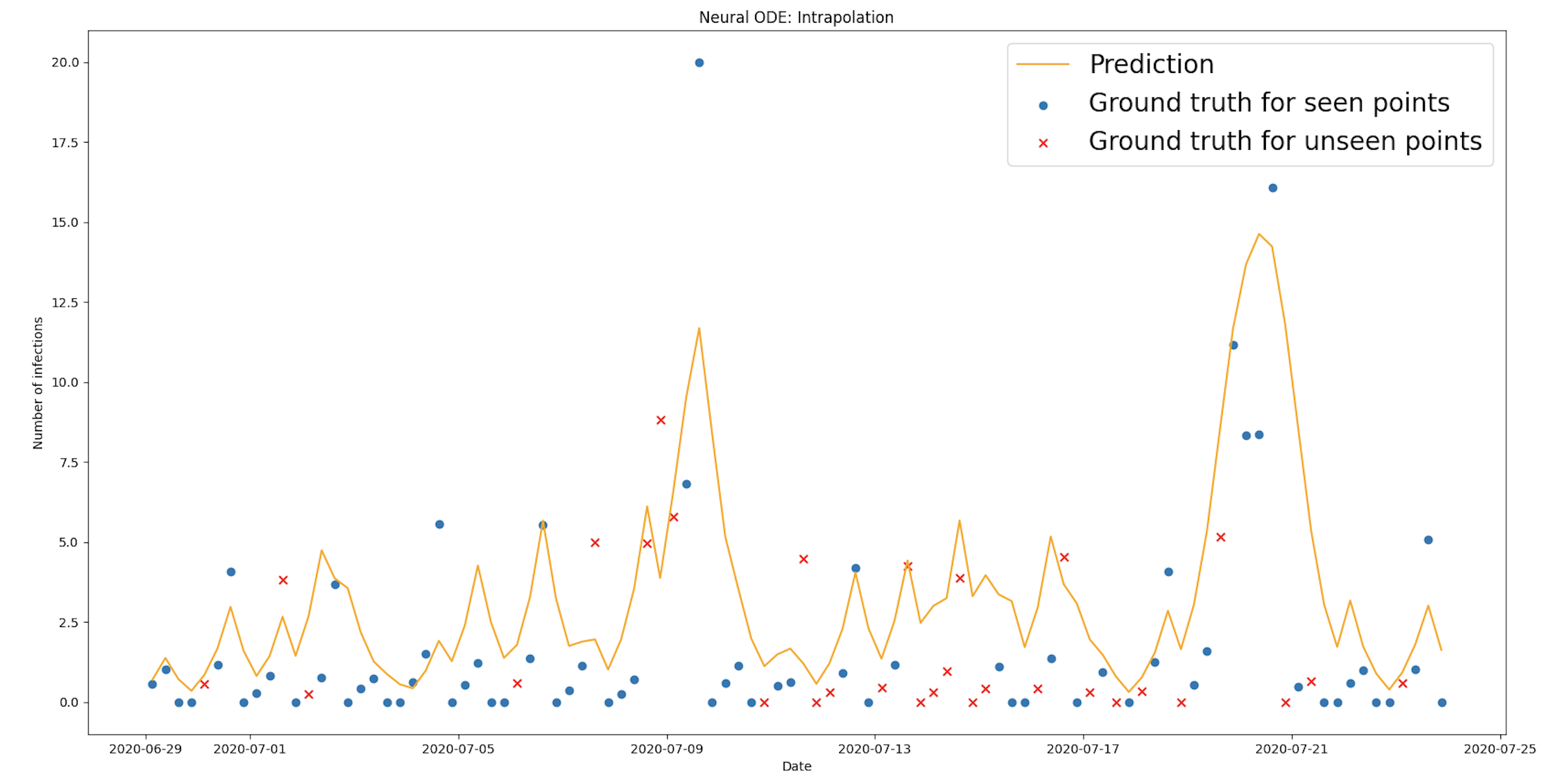}
    \caption{dropping rate $= 0.3$}
    \label{fig:0.3}
    \end{subfigure}
    \begin{subfigure}{0.49\textwidth}
    \includegraphics[width=1\textwidth]{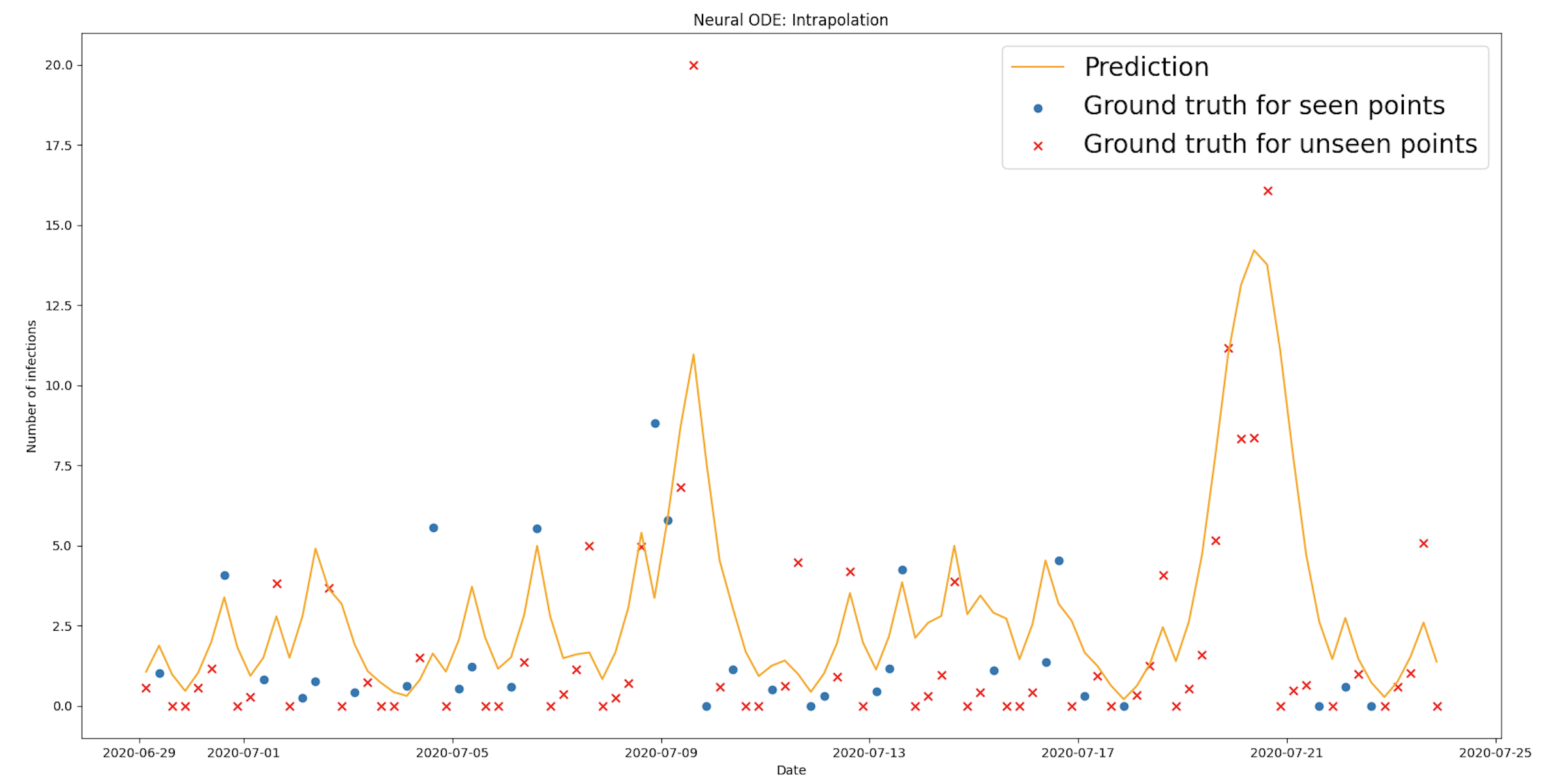}
    \caption{dropping rate $= 0.7$}
    \label{fig:0.7}
    \end{subfigure}
    \begin{subfigure}{0.49\textwidth}
    \includegraphics[width=1\textwidth]{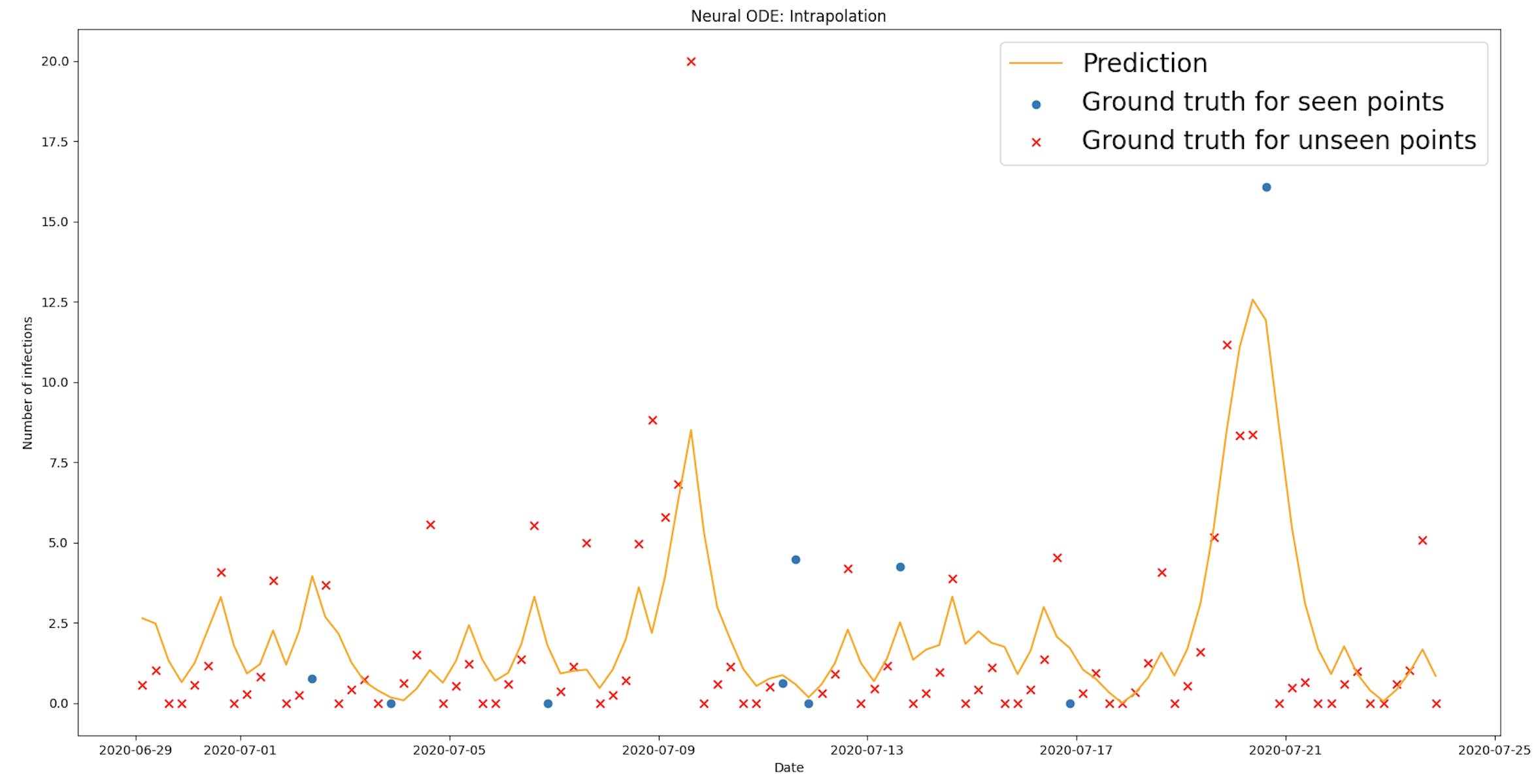}
    \caption{dropping rate $= 0.9$}
    \label{fig:0.9}
    \end{subfigure}
    \caption{Interpolated data windows for MR. NODE with different dropping rate on data points (June 29 2020 - Jul. 25 2020). The model is trained on Costa Rica data.}
    \label{fig:intra}
\end{figure}

To showcase generalization capacities of MR. NODE to unseen irregularities, we tested our model for interpolation at irregular times. When encoding an data window of size 100, we randomly dropped data points with rates $p = \{0.3, 0.7, 0.9\}$ (Table \ref{table:setting}). Then, we interpolated the disease risk at time points either seen or unseen by the model for this window. We then plotted the interpolated windows against the ground truth, where two different markers are used for seen and unseen ground truth points (Figure \ref{fig:intra}).

\section{Discussion}
Our method MR. NODE showed remarkable advantages over RNNs and LSTMs. Firstly, it consistently achieves the lowest extrapolation errors, even as the irregularity in encoded data windows increases (Table \ref{table:results1}). It also outperforms the baselines when trained on two regions' data.

(Figure \ref{fig:intra}) shows MR. NODE's strong capacity at inferring irregular data points. Observing only $30\%$ and $10\%$ of the data respectively (Figure \ref{fig:0.7}, \ref{fig:0.9}), the model still predicts very similar trends as when observing the full data windows (Figure \ref{fig:0}).

In practice, missing values frequently appear in agriculture datasets, especially those recorded in developing countries. Thus, our model's ability to handle irregularity adds great value to our task of predicting black Sigatoka's infection risks in banana plantations.

\section{Future Work}
Since we fixed our extrapolation window size during experiments, we envision future researchers to extend the extrapolation windows of the model and to develop an alert system for peaks in the number of black Sigatoka infections. Reasonable thresholds for Y can be set to give different types of alerts, thus helping farmers to manage the disease and reduce widespread infections. Furthermore, time series with multiple external predictors exist in many other fields. For instance, medical practitioners can apply our model to predict disease onsets for patients.

\section{Conclusion}
We propose a new architecture Multiple predictoR Neural ODEs (MR. NODE), which learnt the dynamics of infections of the black Sigatoka disease directly from data. Successfully modelling time series with multiple predictors, our method enlarged the problem space that latent Neural ODEs can solve. We conducted experiments using semi-real toy datasets and showed our method’s outstanding generalization capacities in forecasting peaks of black Sigatoka infections up to 37.5 days into the future. Importantly, if trained on real good-quality data relevant to the disease, our model can help farmers combat black Sigatoka with preventative actions, thus reducing the cost of banana crop production.

\subsubsection*{Acknowledgments}
This paper and the research behind it would not have been possible without Haotian Cui. He provided the fabulous idea of generating synthetic datasets using statistical methods. We thank William Navarre for offering the idea of studying infectious disease of plants and Ricardo Barros Lourenço for confirming the lack of public crop disease datasets. We thank Zhiyong Dou and Ricky T. Q. Chen for helpful discussions.

\bibliographystyle{unsrt}
\bibliography{main}

\end{document}